\documentclass[conference]{IEEEtran}
\IEEEoverridecommandlockouts

\usepackage{cite}
\usepackage{amsmath,amssymb,amsfonts}
\usepackage{algorithmic}
\usepackage{graphicx}
\usepackage{textcomp}
\usepackage{xcolor}
\usepackage{colortbl}
\def\BibTeX{{\rm B\kern-.05em{\sc i\kern-.025em b}\kern-.08em
    T\kern-.1667em\lower.7ex\hbox{E}\kern-.125emX}}

\usepackage{amssymb}
\usepackage{pifont}
\usepackage{orcidlink}
\usepackage{enumitem}
\setlist{nosep}
\usepackage{multirow}
\usepackage{booktabs}
\usepackage{subcaption}
\usepackage{hyperref}
\usepackage[normalem]{ulem}
\useunder{\uline}{\ul}{}
\hypersetup{
    colorlinks=true,
    linkcolor=blue,
    filecolor=magenta,      
    urlcolor=cyan,
}
\definecolor{lightgray}{gray}{0.9}

\newcommand{\cmark}{\ding{51}}%
\newcommand{\xmark}{\ding{55}}%

\usepackage{xspace}
\makeatletter
\DeclareRobustCommand\onedot{\futurelet\@let@token\@onedot}
\def\@onedot{\ifx\@let@token.\else.\null\fi\xspace}

\def\eg{\emph{e.g}\onedot}

\begin{document}

\title{Rethinking Text-to-Image as Semantic-Aware Data Augmentation for Indoor Scene Recognition}


\author{
\IEEEauthorblockN{
Trong-Vu Hoang\textsuperscript{\rm 1, 2}\orcidlink{0000-0001-7367-1401}, Quang-Binh Nguyen\textsuperscript{\rm 1, 2}\orcidlink{0000-0003-1199-3661}, Dinh-Khoi Vo\textsuperscript{\rm 1, 2}\orcidlink{0000-0001-8831-8846}, Hoai-Danh Vo\textsuperscript{\rm 1, 2}, \\Minh-Triet Tran\textsuperscript{\rm 1, 2}\orcidlink{0000-0003-3046-3041}, Trung-Nghia Le\textsuperscript{\rm 1, 2, *}\orcidlink{0000-0002-7363-2610}\thanks{*Corresponding author. Email address: ltnghia@fit.hcmus.edu.vn}\\
}

\IEEEauthorblockA{%
\textsuperscript{\rm 1}
\textit{University of Science, VNU-HCM, Vietnam}\\%
\textsuperscript{\rm 2}
\textit{Vietnam National University, Ho Chi Minh City, Vietnam}\\%
\ }
}

\maketitle

\begin{abstract}

In the realm of computer vision, indoor image recognition presents challenges due to the intricate interplay of lighting conditions, occlusions, and diverse object arrangements within confined spaces. To address the lacks of training indoor images, we introduce a novel approach leveraging Stable Diffusion (SD) for the generation of synthetic images, which serve as a powerful data augmentation tool. The utilization of SD offers a principled framework for synthesizing diverse and realistic indoor scenes, thereby enriching the training data pool for robust indoor image recognition models. Experimental findings on the MIT Indoor Scene dataset reveal the potential of our proposed approach in enhancing the training of deep models when authentic data is limited. Furthermore, to prevent the misuse of SD synthetic images, we introduce a counter measure based on DIffusion Reconstruction Error (DIRE). The powerful DIRE presentation enables training robust classifiers only using lightweight deep models. Experiments show that our approach can perfectly recognize SD generated images with the accuracy of 100\% using MobilenetV3.
\end{abstract}


\section{Introduction}

In recent years, the advancements in artificial intelligence (AI) have significantly propelled the boundaries of computer vision, enabling machines to perceive and understand visual information with unprecedented accuracy and efficiency. Among various domains within computer vision, indoor image recognition stands out as a crucial component powering various domains ranging from robotics and augmented reality to smart home systems and security surveillance~\cite{gupta2013perceptual,espinace2010indoor,afif2020indoor,afif2020deep}. However, the effectiveness of indoor image recognition models~\cite{quattoni2009recognizing} heavily relies on the quality and diversity of training data, posing a significant challenge in scenarios where collecting large and diverse datasets is impractical or costly.

Data augmentation techniques~\cite{shorten2019survey,wei2019eda,perez2017effectiveness,van2001art,mikolajczyk2018data} have emerged as a promising approach to mitigate the limitations imposed by the scarcity of annotated training data. By synthesizing additional training samples through transformations such as rotation, translation, and scaling, data augmentation techniques aim to enrich the diversity and robustness of the training dataset. Nevertheless, conventional data augmentation methods often struggle to faithfully capture the intricate characteristics of indoor scenes, particularly in scenarios involving occlusions, reflections, and diverse object arrangements.

Generative AI (\eg, generative adversarial networks (GANs)~\cite{Goodfellow-GAN2020} and diffusion models~\cite{rombach2022high}) continues to advance significantly in image generation. Numerous robust models have emerged, capable of generating images based on textual descriptions~\cite{ramesh2022hierarchical, imagegen, kang2023scaling, tao2022df, tao2023galip}. Notably, diffusion models~\cite{rombach2022high, ramesh2022hierarchical, imagegen} have garnered attention for their ability to produce images of remarkable quality and realism. Among these techniques, Stable Diffusion (SD)~\cite{rombach2022high} 
utilizes the power of large-language models (LLMs) for high-quality text-to-image generation across various domains.

In this paper, we introduce a novel approach leveraging SD to address the shortcomings of conventional data augmentation techniques in the context of indoor image recognition. We focus on harnessing the capabilities of SD to generate synthetic indoor images that closely resemble real-world scenes, thereby enriching the diversity and complexity of the training dataset with variations in lighting, textures, and object configurations. We also propose a duplicated and outlier removal module to eliminate images which resemble to others. This ensures the coherence among generated images for enhancing the accuracy and reliability of training models.

Our approach presents several undeniable advantages. SD has ability to synthesize diverse indoor scenes with realistic textures, lighting conditions, and spatial configurations, effectively capturing the inherent variability present in real-world environments. Furthermore, by incorporating strength of LLMs, SD can steer the image generation process towards producing semantically relevant variations, aligning closely with the task of indoor image recognition. We demonstrate the effectiveness of our proposed approach through extensive experiments on the well-known MIT Indoor Scene dataset~\cite{quattoni2009recognizing} (MIT in short). By incorporating synthetically generated images as part of the training data, we showcase substantial improvements in the performance of image recognition models. A simple EfficientNetV2~\cite{tan2021efficientnetv2} trained on our generated augmented images can achieve the accuracy of $84.2\%$. Therefore, our augmentation strategy to suit the complexity of indoor image recognition tasks. 

\begin{figure*}[t!]
	\centering
	\includegraphics[width=0.9\textwidth]{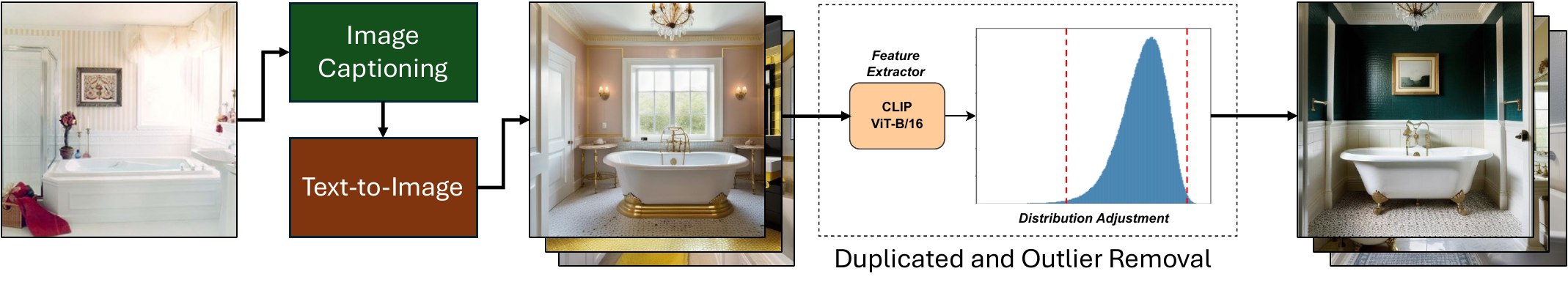}
	\caption{The dataset synthesis process consists of three key phases: prompt generation, image generation, and duplicated \& outlier removal.}
	\label{fig:overall_data_gen}
 \vspace{-3mm}
\end{figure*}

While SD shows advancements as data augmentation, it also introduces risks, particularly when generated images are misused or repurposed for malicious intents~\cite{ovadya2019reducing}. In the wrong hands, such images could be employed to deceive, manipulate, or exploit individuals and systems. It poses significant ethical and security concerns~\cite{solaiman2023evaluating}. To address these risks and safeguard against potential misuse, it becomes imperative to develop proactive measures that enable the detection of images generated by SD.

In this paper, we train deep learning models specifically to recognize and distinguish between authentic and artificially generated images. Particularly, we utilize DIffusion Reconstruction Error (DIRE)~\cite{wang2023dire}, which measures the error between an input image and its reconstruction counterpart by a pre-trained diffusion model. Leveraging the inherent patterns and discrepancies inherent of artificial images in DIRE presentation, we can train robust classifiers only using lightweight deep models. 
Our experimental results demonstrate the capability of our proposed approach to accurately identify generated images with a remarkable rate of 100\% using the lightweight MobilenetV3~\cite{howard2019searching}. Our method can serve as a frontline defense against potential misuse, enabling the identification of synthetic content with high accuracy and reliability.

In summary, our main contributions are as follows:
\begin{itemize}
    \item We introduce an approach that empowers training image recognition models by using SD to create diverse indoor images as the augmented data. We also showcase the real-world applicability of our generated data in deep learning model training, highlighting the effectiveness of our method in providing deep models for a wide range of image recognition tasks.

    \item We present an effective defense against potential misuse of SD synthetic images. By effectively discerning between authentic and generated images via DIRE presentation, we not only reduces number of training parameters of deep models but also keep remarkable accuracy rate of 100\% using MobilenetV3. 
\end{itemize}

The structure of this paper is as follows. Section \ref{sec:related_work} discusses previous works. Our proposed method is presented in Section \ref{sec:proposed_method}. Section \ref{sec:experiments} shows experimental results and discussions. Finally, in Section \ref{sec:conclusion}, we summarize the paper.

\begin{figure*}[t!]
	\centering
	\includegraphics[width=0.8\textwidth]{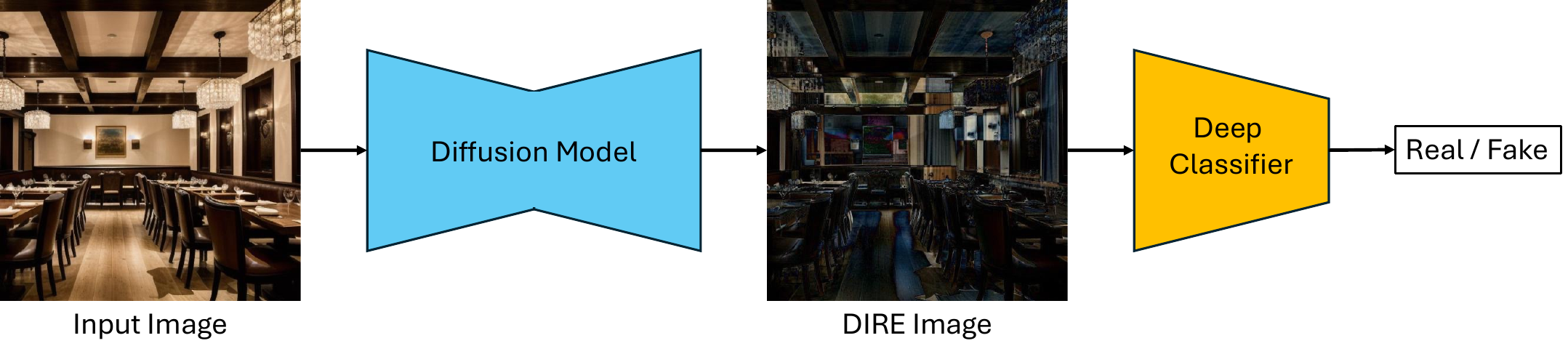}
	\caption{SD generated image recognition process includes two phases: DIRE image creation and fake image recognition.}
	\label{fig:dire}
 \vspace{-3mm}
\end{figure*}

\section{Related Work}
\label{sec:related_work}

\subsection{Data Augmentation for Image Classification}

Small datasets often lead to models that struggle with generalization, resulting in overfitting issues. To address this challenge, various data augmentation methods have been developed. Data augmentation techniques, which increase the amount of training data, play a pivotal role in enhancing the robustness and generalization capabilities of deep models. Common augmentation methods include random rotations, translations, flips, zooms, and changes in brightness or contrast. By applying such transformations to the training images, data augmentation mitigates overfitting and equips the model with a more comprehensive understanding of the underlying features, leading to improved classification accuracy and stability.

Furthermore, recent advancements in data augmentation have extended beyond traditional geometric transformations to include more sophisticated techniques such as Mixup~\cite{zhang2017mixup} and CutMix~\cite{yun2019cutmix}. These methods involve blending or replacing patches of images, interpolating between multiple images to introduce additional variability and encourage the model to learn more robust representations.

Different from previous work, we utilize SD to generate high-quality and diverse synthetic images using extracted textual information from the training set. This not only increases the size of the training set but also enhances its diversity, enabling deep models to learn more robust representations and improve generalization performance.



\subsection{Text-to-Image}

With the introduction of large-scale datasets~\cite{schuhmann2022laion}, generative model-based text-to-image~\cite{saharia2022photorealistic,ramesh2021zero} has experienced rapid advancements, garnering substantial attention from both the scientific community and the general public. Notably, these advancements centre around two dominant approaches: GANs~\cite{sauer2023stylegan, liao2022text, kang2023scaling} and diffusion models~\cite{rombach2022high, saharia2022photorealistic, balaji2022ediffi, nichol2021glide, ramesh2022hierarchical}. GAN-based methods known for high-fidelity image synthesis and fast inference but they often grapple with challenges related to training instability and mode collapse, as well as the diversity of generated images. On the other hand, diffusion-based models have emerged as an exceptional capacity for generating high-quality and realistic images.

Pioneering works~\cite{nichol2021glide, ramesh2022hierarchical} integrated diffusion model with classifier guidance to improve image quality. However, these models do not achieve a good balance between efficiency and image fidelity. After that, to balance efficiency and fidelity, novel techniques have emerged, such as the coarse-to-fine generation~\cite{saharia2022photorealistic, balaji2022ediffi} and latent space exploration~\cite{rombach2022high}. Built on the latent diffusion concept, SD~\cite{rombach2022high} 
stood out as the first open-source large model, significantly expanding text-to-image synthesis capabilities. 

In this paper, we adopt SD for the text-to-image generation module, drawn to its remarkable image quality. The availability of robust pre-trained weights also underscores the versatility and impact of SD in advancing text-to-image generation.

\section{Proposed Method}
\label{sec:proposed_method}

\subsection{Augmented Data Generation}

Our augmented data generation process involves three phases: prompt generation, image generation, and duplicated and outlier removal, as illustrated in Fig.~\ref{fig:overall_data_gen}. First of all, the prompt generation phase involves an image captioning model to create the image description. The prompts are then fed into a text-to-image model to generate corresponding images in the image generation phase. Finally, to ensure the generated images’ quality, the duplicated and outlier removal phase filters out all the unqualified images.

\subsubsection{Prompt Generation}

The quality and realisticness of images generated by text-to-image models heavily relies on input prompts. To obtain high-quality prompts, our methodology involves using the CLIP~\cite{clip} text decoder to extract textual prompts directly from images to ensure the contextual fidelity of each image to its class. This methodology is deliberately chosen to avoid the unintended generalization of context across all classes that may occur if prompts are generated randomly directly from only class names. Hence, our approach can maintain generated images' diversity and offer a fair, contextually accurate environment for training deep models.

\subsubsection{Image Generation}

Generated prompts are fed into SD, a cutting-edge text-to-image model, to produce relevant images. The growing interest in generative AI models has led to the emergence of powerful checkpoints for impressive image generation using SD in various domains. Among these, we specifically utilize the Realistic Vision checkpoint due to its ability to produce exceptionally strikingly realistic images. Additionally, we employ techniques like CPU offloading and attention layer acceleration~\cite{xFormers} to optimize memory and time consumption during the image generation process.

\subsubsection{Duplicated and Outlier Removal}

We discover that from the same prompt, we can get some generated images that look significantly similar or different from each other. Upon examining the generated images stemming from the same prompt, we observe variations in their resemblance to one another. This discovery instigates the duplicated and outlier removal phase aimed at ensuring the coherence among generated images within the dataset. This process is crucial for enhancing the accuracy and reliability of subsequent data-driven models and is achieved through image distribution adjustment operation (see Fig.~\ref{fig:overall_data_gen}).

To facilitate the distribution adjustment process, we compute feature vectors for the generated images using CLIP~\cite{clip}, a robust vision-language model, specifically the B-L/16 version. Each generated image is associated with a corresponding feature vector comprising 512 dimensions. Afterwards, in the distribution adjustment process, we assess the image similarity of every image pair using the cosine similarity metric between their extracted features. Let $\bold{u}_i$ be the feature vector of image $i$. The similarity score between a pair of images $i$ and $j$ is denoted as: 
\begin{equation}
    \text{sim}(\bold{u}_i, \bold{u}_j) = \frac{\bold{u}_i \cdot \bold{u}_j}{\left \| \bold{u}_i \right \| \left \| \bold{u}_j \right \|}.
\end{equation}

We also define the number of neighbors of an image $i$ as the number of image $j$ ($j \neq i$) by considering their similarity scores in the range $[\alpha, \beta]$. Our aim is to ensure that the similarity score between any pair of generated images falls within the acceptable range $[\alpha, \beta]$. If a pair fails to meet this condition, the image with fewer neighbors is eliminated. This process effectively eliminates duplicates (where scores are greater than $\beta$) and irrelevant images (where scores are less than $\alpha$). We empirically set $\beta=0.9825$, retaining ratio of $0.9$.

\begin{figure*}[t!]
    \centering
\includegraphics[width=\linewidth]{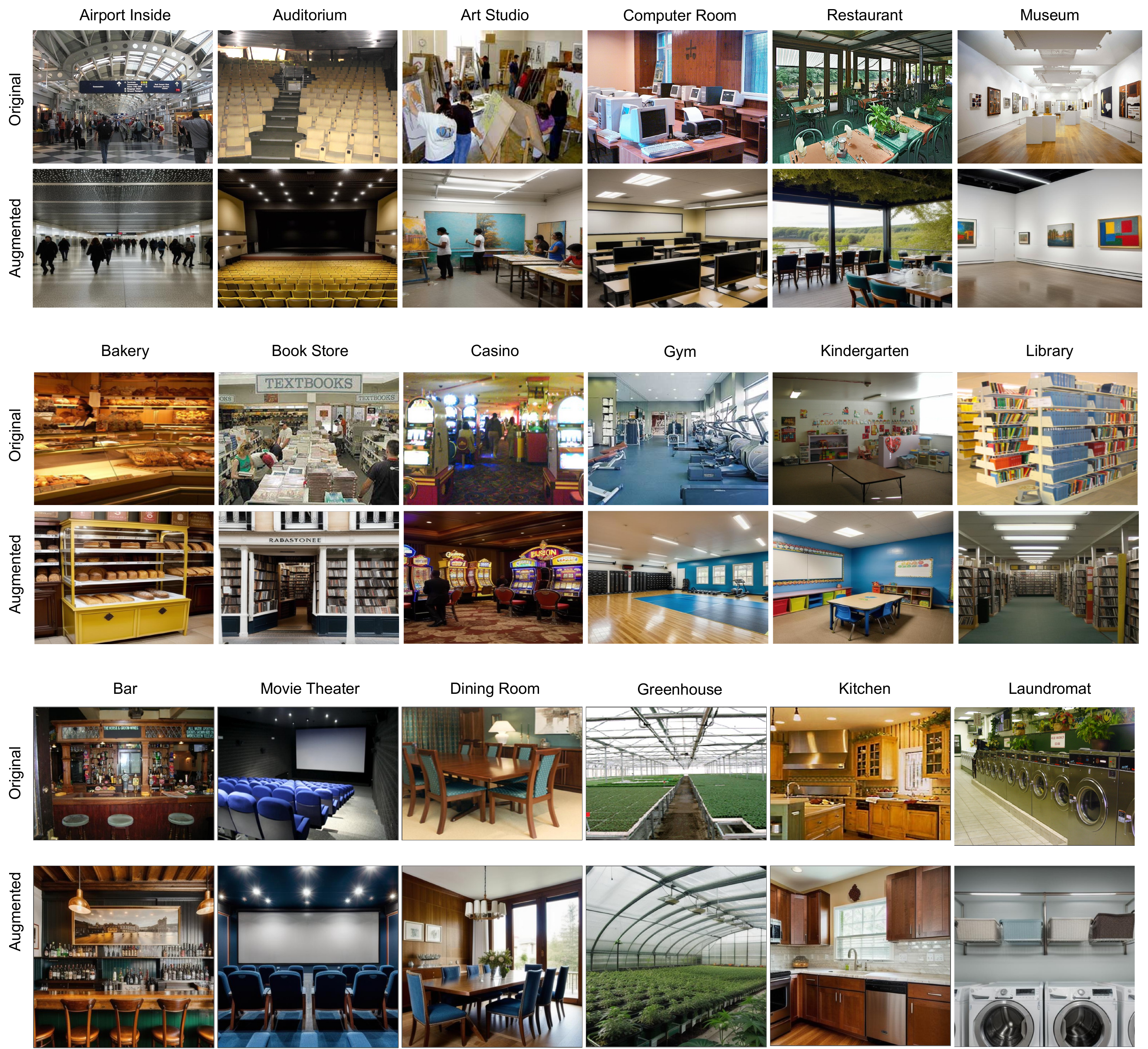}
    \caption{Examples of original images in MIT dataset~\cite{quattoni2009recognizing} (top) and our augmented images (bottom).}
    \label{fig:simulated-dataset}
\end{figure*}

\begin{figure}[t!]
    \centering
    \includegraphics[width=\linewidth]{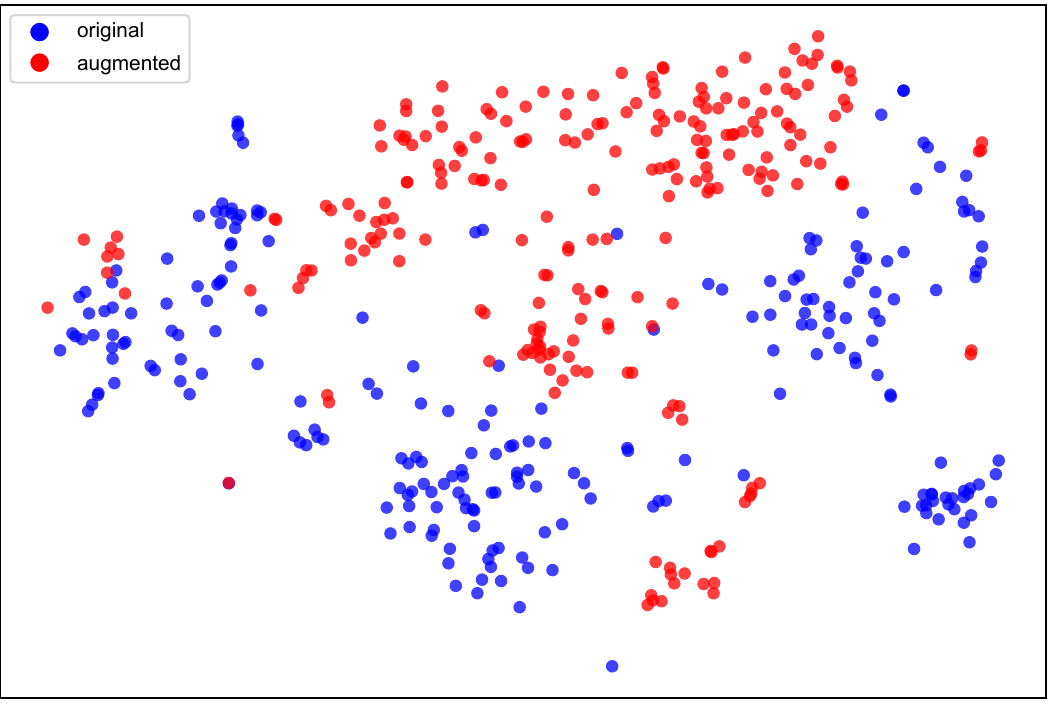}
    \caption{Feature space visualization of original and augmented images in MIT dataset~\cite{quattoni2009recognizing}.}
    \label{fig:feature-space-mit}
\end{figure}

\subsection{SD Generated Image Recognition}

While SD synthetic images  have useful applications, there are concerns about privacy and ethical problems if the images are used for malicious intents. To address these risks, it is an urgent demand for SD generated image detector. A naïve approach is to directly train CNN classifiers on RGB real images and SD generated images. However, such a naïve scheme suffers limited generalization, and we may need to train a very large model to achieve a satisfactory performance. 

Among various image representations available, we notice that DIffusion Reconstruction Error (DIRE)~\cite{wang2023dire}, measuring reconstruction errors of images inverted and reconstructed by DDIM~\cite{mokady2023null}, provides a general image representation. We argue that DIRE images can efficiently help train robust classifiers with less learned parameters. We remark that combining RGB with DIRE hurts generalization compared to pure DIRE~\cite{wang2023dire}. Hence, we use only DIRE images for training binary deep classifiers  as shown in Fig.~\ref{fig:dire}.

\section{Experiments}
\label{sec:experiments}

\subsection{Implementation Details}

In our experiments, EfficientNetV2~\cite{tan2021efficientnetv2}, MobilenetV3~\cite{howard2019searching}, Resnet50~\cite{he2016deep}, ViT-small~\cite{dosovitskiy2020image}, Swin-tiny~\cite{liu2021swin}, pre-trained on ImageNet~\cite{deng2009imagenet}, was used. The networks were fine-tuned for 20 epochs. We used a learning rate of $10^{-3}$, a batch size of 128, the AdamW optimizer~\cite{loshchilov2017decoupled}, and employed Sparse Categorical Cross-entropy~\cite{zhang2018generalized} as the loss function.

\subsection{Experimental Settings}

We used a widely recognized benchmark for indoor scene image recognition, the MIT Indoor Scene dataset~\cite{quattoni2009recognizing} (MIT in short), in our experiments. This dataset encompasses 67 categories of indoor scene images. There are 15,620 images in total, with at least 100 images in each category. We follow the division of training set and test set in~\cite{quattoni2009recognizing} with 80 images of each category are used for training, and 20 images are used for testing.


\subsection{Augmented Image Generation and Analysis}

Our approach centers on the creation of augmented images that effectively encapsulates the core characteristics of the MIT dataset. Using the proposed framework, we obtained a augmented images derived from the MIT dataset, mirroring the number of images present in the original training set. Figure~\ref{fig:simulated-dataset} presents a side-by-side comparison of images from the MIT dataset and their corresponding replicated versions. 

To confirm the authenticity of the augmented images, we employed a feature space visualization approach with CLIP image features~\cite{clip} and t-SNE dimension reduction~\cite{van2008visualizing}. In this analysis, we randomly selected three classes from the MIT dataset and sampled images to represent both the original and augmented sets. The results of this feature visualization are showcased in Fig.~\ref{fig:feature-space-mit}. This figure reveals two clusters, one for the original images and one for the augmented images. Notably, these clusters exhibit nearly overlapping distributions, indicating that the augmented images can preserve the essential characteristics of the MIT dataset, indicating a high degree of fidelity in the augmented images.

\subsection{Effectiveness of Data Augmentation}

This experiment aims to leverage generated images as valuable augmented data to enhance deep model training, in conjunction with the original dataset. To accomplish this, the generated images were merged into the original images with the ratio 1:1 to train deep classifiers. These trained models were then rigorously evaluated using the original test set. In this experiment, we used EfficientNetV2 as the classifier. Table~\ref{tab:evaluation-results} showcases the performance in terms of accuracy on the test set, which achieves 84\%, improving up to 0.7\% compared with not using our data augmentation approach. The results underscore the potential advantages of integrating generated data into the training process. 

We also verify the effectiveness of our augmented images in the context where real data is limited or difficult to obtain. Our evaluation process involves random sampling from both the 50\% real images and generated images to train the EfficientNetV2 classifier. It is worth mentioning that our evaluation process involves five times of random sampling to ensure the robustness of trained models. The results are then averaged across these iterations to provide a comprehensive perspective on the impact of incorporating generated data. As seen in Table~\ref{tab:evaluation-results}, averaged results demonstrate that the performance of the deep model using augmented images surpasses that of the model trained solely on real images up to 1.9\% in accuracy. These findings emphasizes the practical implication of our data augmentation approach in which generated data can substantially enhance the training process by serving as an additional data source, particularly when acquiring real data is limited or challenging.

\begin{table}[t!]
    \centering
    \caption{Evaluation results of using data augmentation.}
    \label{tab:evaluation-results}
    \begin{tabular}{|c|c|c|}
        \hline
        \textbf{\#Real Images} & \textbf{Use Augmented Images} & \textbf{Accuracy (\%)} \\
        \hline
         5,200 & \xmark & 83.5 
        \\
        \rowcolor{lightgray} \textbf{5,200} & \cmark &  \textbf{84.2}
        \\
        \hline
        2,600 & \xmark & 71.9
         \\
        \rowcolor{lightgray} \textbf{2,600} & \cmark & \textbf{73.8}
         \\ 
        \hline
    \end{tabular}
\end{table}

\subsection{Generated Image Recognition Evaluation}

\begin{table}[t!]
    \centering
    \caption{Performance of different classifiers in identifying SD synthetic images.}
    \label{tab:dire_eval}
    \resizebox{\linewidth}{!}{
    \begin{tabular}{|c|c|r|c|}
        \hline
        \textbf{Method} & \textbf{Train} & \textbf{Trainable} & \textbf{Accuracy (\%)} \\
        \textbf{} & \textbf{All Layers} & \textbf{Parameters} & \textbf{} \\
        \hline
        Resnet50 & \xmark & 4098 & 94.0 \\
        EfficientNetV2 & \xmark & 3586 & 89.5 \\
        MobilenetV3 & \xmark & 2562 & 91.2 \\
        ViT-small & \xmark & 770  & 90.5  \\
        Swin-tiny & \xmark & 1538  & 91.2  \\
        \rowcolor{lightgray} \textbf{MobilenetV3 + DIRE} & \xmark & \textbf{2562} & \textbf{97.2} \\
        \hline
        Resnet50 & \cmark & 23.5M & 99.2 \\
        EfficientNetV2 & \cmark & 22.1M & 99.6 \\
        MobilenetV3 & \cmark & 4M & 99.9 \\
        Swin-tiny & \cmark & 27.5M  & 99.1  \\
        ViT-small & \cmark & 22.4M  & 96.3  \\
        \rowcolor{lightgray} \textbf{MobilenetV3 + DIRE} & \cmark & \textbf{4M} & \textbf{100} \\ 
        \hline
    \end{tabular}
    }
\end{table}

We examine various deep classifiers (\eg, EfficientNetV2~\cite{tan2021efficientnetv2}, MobilenetV3~\cite{howard2019searching}, Resnet50~\cite{he2016deep}, ViT-small~\cite{dosovitskiy2020image}, and Swin-tiny~\cite{liu2021swin}) in recognizing Stable Diffusion generated images. Through experimentation across different settings, such as utilizing RGB versus DIRE images and employing different fine-tuning strategies, we observe a significant improvement in accuracy when utilizing DIRE images. Notably, when fine-tuning only the last layer on DIRE images, MobilenetV3 outperformed other methods by a substantial margin ranging from 3\% to 8\% (see Table~\ref{tab:dire_eval}). This superiority, coupled with the advantage of training only 2500 parameters, underscores MobilenetV3's efficiency and effectiveness in capturing patterns within SD generated images. 

Furthermore, our experiments demonstrate MobilenetV3's robustness. Despite its compact architecture with 4 million parameters, MobilenetV3 demonstrates remarkable efficacy, achieving a perfect 100\% accuracy compared to larger deep networks exceeding 20 million parameters. This robust performance highlights MobilenetV3 as a compelling choice for tasks requiring both accuracy and computational efficiency in recognizing synthetic imagery.

\section{Conclusion}
\label{sec:conclusion}

This paper emphasizes the potential of SD as a valuable tool for enhancing indoor image recognition through data augmentation. Through empirical evaluations and qualitative analyses, we illustrate the effectiveness of our proposed approach in addressing the data scarcity and variability issues inherent in indoor image recognition tasks. We believe that our work not only advances the state-of-the-art in indoor image recognition but also opens up new avenues for leveraging generative modeling techniques to augment training datasets and improve the performance of computer vision systems across diverse application domains.

On the other hand, to prevent the misuse of SD synthetic images, we introduce a defense approach using DIRE to training a lightweight yet robust classifier. Experimental results indicate that our approach can recognize generated images with the accuracy of 100\%.

\section*{Acknowledgement}

This research is funded by University of Science, VNU-HCM, under grant number CNTT 2023-04.

\bibliographystyle{IEEEtran}
\bibliography{short_bibtex}

@Misc{xFormers,
  author =       {Benjamin Lefaudeux and Francisco Massa and Diana Liskovich and Wenhan Xiong and Vittorio Caggiano and Sean Naren and Min Xu and Jieru Hu and Marta Tintore and Susan Zhang and Patrick Labatut and Daniel Haziza},
  title =        {xFormers: A modular and hackable Transformer modelling library},
  howpublished = {\url{https://github.com/facebookresearch/xformers}},
  year =         {2022}
}

@inproceedings{kang2023scaling,
  title={Scaling up gans for text-to-image synthesis},
  author={Kang, Minguk and Zhu, Jun-Yan and Zhang, Richard and Park, Jaesik and Shechtman, Eli and Paris, Sylvain and Park, Taesung},
  booktitle={CVPR},
  pages={10124--10134},
  year={2023}
}

@inproceedings{liao2022text,
  title={Text to image generation with semantic-spatial aware gan},
  author={Liao, Wentong and Hu, Kai and Yang, Michael Ying and Rosenhahn, Bodo},
  booktitle={CVPR},
  pages={18187--18196},
  year={2022}
}

@inproceedings{he2016deep,
  title={Deep residual learning for image recognition},
  author={He, Kaiming and Zhang, Xiangyu and Ren, Shaoqing and Sun, Jian},
  booktitle={CVPR},
  pages={770--778},
  year={2016}
}

@article{schuhmann2022laion,
  title={Laion-5b: An open large-scale dataset for training next generation image-text models},
  author={Schuhmann, Christoph and Beaumont, Romain and Vencu, Richard and Gordon, Cade and Wightman, Ross and Cherti, Mehdi and Coombes, Theo and Katta, Aarush and Mullis, Clayton and Wortsman, Mitchell and others},
  journal={NeurIPS},
  volume={35},
  pages={25278--25294},
  year={2022}
}

@article{sauer2023stylegan,
  title={Stylegan-t: Unlocking the power of gans for fast large-scale text-to-image synthesis},
  author={Sauer, Axel and Karras, Tero and Laine, Samuli and Geiger, Andreas and Aila, Timo},
  journal={arXiv preprint arXiv:2301.09515},
  year={2023}
}

@inproceedings{rombach2022high,
  title={High-resolution image synthesis with latent diffusion models},
  author={Rombach, Robin and Blattmann, Andreas and Lorenz, Dominik and Esser, Patrick and Ommer, Bj{\"o}rn},
  booktitle={CVPR},
  pages={10684--10695},
  year={2022}
}

@article{saharia2022photorealistic,
  title={Photorealistic text-to-image diffusion models with deep language understanding},
  author={Saharia, Chitwan and Chan, William and Saxena, Saurabh and Li, Lala and Whang, Jay and Denton, Emily L and Ghasemipour, Kamyar and Gontijo Lopes, Raphael and Karagol Ayan, Burcu and Salimans, Tim and others},
  journal={NeurIPS},
  volume={35},
  pages={36479--36494},
  year={2022}
}

@article{balaji2022ediffi,
  title={ediffi: Text-to-image diffusion models with an ensemble of expert denoisers},
  author={Balaji, Yogesh and Nah, Seungjun and Huang, Xun and Vahdat, Arash and Song, Jiaming and Kreis, Karsten and Aittala, Miika and Aila, Timo and Laine, Samuli and Catanzaro, Bryan and others},
  journal={arXiv preprint arXiv:2211.01324},
  year={2022}
}

@article{nichol2021glide,
  title={Glide: Towards photorealistic image generation and editing with text-guided diffusion models},
  author={Nichol, Alex and Dhariwal, Prafulla and Ramesh, Aditya and Shyam, Pranav and Mishkin, Pamela and McGrew, Bob and Sutskever, Ilya and Chen, Mark},
  journal={arXiv preprint arXiv:2112.10741},
  year={2021}
}

@article{ramesh2022hierarchical,
  title={Hierarchical text-conditional image generation with clip latents},
  author={Ramesh, Aditya and Dhariwal, Prafulla and Nichol, Alex and Chu, Casey and Chen, Mark},
  journal={arXiv preprint arXiv:2204.06125},
  volume={1},
  number={2},
  pages={3},
  year={2022}
}

@article{van2008visualizing,
  title={Visualizing data using t-SNE.},
  author={Van der Maaten, Laurens and Hinton, Geoffrey},
  journal={Journal of machine learning research},
  volume={9},
  number={11},
  year={2008}
}

@article{imagegen,
  title={Photorealistic text-to-image diffusion models with deep language understanding},
  author={Saharia, Chitwan and Chan, William and Saxena, Saurabh and Li, Lala and Whang, Jay and Denton, Emily L and Ghasemipour, Kamyar and Gontijo Lopes, Raphael and Karagol Ayan, Burcu and Salimans, Tim and others},
  journal={NeurIPS},
  volume={35},
  pages={36479--36494},
  year={2022}
}

@inproceedings{clip,
  title={Learning transferable visual models from natural language supervision},
  author={Radford, Alec and Kim, Jong Wook and Hallacy, Chris and Ramesh, Aditya and Goh, Gabriel and Agarwal, Sandhini and Sastry, Girish and Askell, Amanda  and others},
  booktitle={ICML},
  pages={8748--8763},
  year={2021},
  @@organization={PMLR}
}

@inproceedings{quattoni2009recognizing,
  title={Recognizing indoor scenes},
  author={Quattoni, Ariadna and Torralba, Antonio},
  booktitle={CVPR},
  pages={413--420},
  year={2009},
  @@organization={IEEE}
}

@inproceedings{tan2021efficientnetv2,
  title={Efficientnetv2: Smaller models and faster training},
  author={Tan, Mingxing and Le, Quoc},
  booktitle={ICML},
  pages={10096--10106},
  year={2021},
  @@organization={PMLR}
}

@inproceedings{deng2009imagenet,
  title={Imagenet: A large-scale hierarchical image database},
  author={Deng, Jia and Dong, Wei and Socher, Richard and Li, Li-Jia and Li, Kai and Fei-Fei, Li},
  booktitle={CVPR},
  pages={248--255},
  year={2009},
  @@organization={Ieee}
}

@article{loshchilov2017decoupled,
  title={Decoupled weight decay regularization},
  author={Loshchilov, Ilya and Hutter, Frank},
  journal={arXiv preprint arXiv:1711.05101},
  year={2017}
}

@inproceedings{liu2021swin,
  title={Swin transformer: Hierarchical vision transformer using shifted windows},
  author={Liu, Ze and Lin, Yutong and Cao, Yue and Hu, Han and Wei, Yixuan and Zhang, Zheng and Lin, Stephen and Guo, Baining},
  booktitle={CVPR},
  pages={10012--10022},
  year={2021}
}

@article{Goodfellow-GAN2020,
  title={Generative adversarial networks},
  author={Goodfellow, Ian and Pouget-Abadie, Jean and Mirza, Mehdi and Xu, Bing and Warde-Farley, David and Ozair, Sherjil and Courville, Aaron and Bengio, Yoshua},
  journal={Communications of the ACM},
  volume={63},
  number={11},
  pages={139--144},
  year={2020},
}

@inproceedings{tao2022df,
  title={Df-gan: A simple and effective baseline for text-to-image synthesis},
  author={Tao, Ming and Tang, Hao and Wu, Fei and Jing, Xiao-Yuan and Bao, Bing-Kun and Xu, Changsheng},
  booktitle={CVPR},
  pages={16515--16525},
  year={2022}
}

@inproceedings{tao2023galip,
  title={GALIP: Generative Adversarial CLIPs for Text-to-Image Synthesis},
  author={Tao, Ming and Bao, Bing-Kun and Tang, Hao and Xu, Changsheng},
  booktitle={CVPR},
  pages={14214--14223},
  year={2023}
}

@article{shorten2019survey,
  title={A survey on image data augmentation for deep learning},
  author={Shorten, Connor and Khoshgoftaar, Taghi M},
  journal={Journal of big data},
  volume={6},
  number={1},
  pages={1--48},
  year={2019},
  @publisher={Springer}
}

@article{wei2019eda,
  title={Eda: Easy data augmentation techniques for boosting performance on text classification tasks},
  author={Wei, Jason and Zou, Kai},
  journal={arXiv preprint arXiv:1901.11196},
  year={2019}
}

@article{perez2017effectiveness,
  title={The effectiveness of data augmentation in image classification using deep learning},
  author={Perez, Luis and Wang, Jason},
  journal={arXiv preprint arXiv:1712.04621},
  year={2017}
}

@article{van2001art,
  title={The art of data augmentation},
  author={Van Dyk, David A and Meng, Xiao-Li},
  journal={Journal of Computational and Graphical Statistics},
  volume={10},
  number={1},
  pages={1--50},
  year={2001},
  @publisher={Taylor \& Francis}
}

@inproceedings{gupta2013perceptual,
  title={Perceptual @organization and recognition of indoor scenes from RGB-D images},
  author={Gupta, Saurabh and Arbelaez, Pablo and Malik, Jitendra},
  booktitle={CVPR},
  pages={564--571},
  year={2013}
}

@inproceedings{espinace2010indoor,
  title={Indoor scene recognition through object detection},
  author={Espinace, Pablo and Kollar, Thomas and Soto, Alvaro and Roy, Nicholas},
  booktitle={ICRA},
  pages={1406--1413},
  year={2010},
  @organization={IEEE}
}

@inproceedings{mikolajczyk2018data,
  title={Data augmentation for improving deep learning in image classification problem},
  author={Miko{\l}ajczyk, Agnieszka and Grochowski, Micha{\l}},
  booktitle={2018 international interdisciplinary PhD workshop (IIPhDW)},
  pages={117--122},
  year={2018},
  @organization={IEEE}
}

@article{dosovitskiy2020image,
  title={An image is worth 16x16 words: Transformers for image recognition at scale},
  author={Dosovitskiy, Alexey and Beyer, Lucas and Kolesnikov, Alexander and Weissenborn, Dirk and Zhai, Xiaohua and Unterthiner, Thomas and Dehghani, Mostafa and Minderer, Matthias and Heigold, Georg and Gelly, Sylvain and others},
  journal={arXiv preprint arXiv:2010.11929},
  year={2020}
}

@inproceedings{wang2023dire,
  title={Dire for diffusion-generated image detection},
  author={Wang, Zhendong and Bao, Jianmin and Zhou, Wengang and Wang, Weilun and Hu, Hezhen and Chen, Hong and Li, Houqiang},
  booktitle={ICCV},
  pages={22445--22455},
  year={2023}
}

@inproceedings{howard2019searching,
  title={Searching for mobilenetv3},
  author={Howard, Andrew and Sandler, Mark and Chu, Grace and Chen, Liang-Chieh and Chen, Bo and Tan, Mingxing and Wang, Weijun and Zhu, Yukun and Pang, Ruoming and Vasudevan, Vijay and others},
  booktitle={ICCV},
  pages={1314--1324},
  year={2019}
}

@inproceedings{mokady2023null,
  title={Null-text inversion for editing real images using guided diffusion models},
  author={Mokady, Ron and Hertz, Amir and Aberman, Kfir and Pritch, Yael and Cohen-Or, Daniel},
  booktitle={CVPR},
  pages={6038--6047},
  year={2023}
}

@inproceedings{yun2019cutmix,
  title={Cutmix: Regularization strategy to train strong classifiers with localizable features},
  author={Yun, Sangdoo and Han, Dongyoon and Oh, Seong Joon and Chun, Sanghyuk and Choe, Junsuk and Yoo, Youngjoon},
  booktitle={ICCV},
  pages={6023--6032},
  year={2019}
}

@article{zhang2017mixup,
  title={mixup: Beyond empirical risk minimization},
  author={Zhang, Hongyi and Cisse, Moustapha and Dauphin, Yann N and Lopez-Paz, David},
  journal={arXiv preprint arXiv:1710.09412},
  year={2017}
}

@article{zhang2018generalized,
  title={Generalized cross entropy loss for training deep neural networks with noisy labels},
  author={Zhang, Zhilu and Sabuncu, Mert},
  journal={NeurIPS},
  volume={31},
  year={2018}
}

@inproceedings{afif2020indoor,
  title={Indoor image recognition and classification via deep convolutional neural network},
  author={Afif, Mouna and Ayachi, Riadh and Said, Yahia and Pissaloux, Edwige and Atri, Mohamed},
  booktitle={International Conference on Sciences of Electronics, Technologies of Information and Telecommunications (SETIT’18), Vol. 1},
  pages={364--371},
  year={2020},
  @organization={Springer}
}

@article{afif2020deep,
  title={Deep learning based application for indoor scene recognition},
  author={Afif, Mouna and Ayachi, Riadh and Said, Yahia and Atri, Mohamed},
  journal={Neural Processing Letters},
  volume={51},
  pages={2827--2837},
  year={2020},
  @publisher={Springer}
}

@article{ovadya2019reducing,
  title={Reducing malicious use of synthetic media research: Considerations and potential release practices for machine learning},
  author={Ovadya, Aviv and Whittlestone, Jess},
  journal={arXiv preprint arXiv:1907.11274},
  year={2019}
}

@article{solaiman2023evaluating,
  title={Evaluating the social impact of generative ai systems in systems and society},
  author={Solaiman, Irene and Talat, Zeerak and Agnew, William and Ahmad, Lama and Baker, Dylan and Blodgett, Su Lin and Daum{\'e} III, Hal and Dodge, Jesse and Evans, Ellie and Hooker, Sara and others},
  journal={arXiv preprint arXiv:2306.05949},
  year={2023}
}

@inproceedings{ramesh2021zero,
  title={Zero-shot text-to-image generation},
  author={Ramesh, Aditya and Pavlov, Mikhail and Goh, Gabriel and Gray, Scott and Voss, Chelsea and Radford, Alec and Chen, Mark and Sutskever, Ilya},
  booktitle={ICML},
  pages={8821--8831},
  year={2021},
  @organization={Pmlr}
}

\end{document}